\documentclass[11pt]{article}

\usepackage[final]{acl}

\usepackage{times}
\usepackage{latexsym}

\usepackage[T1]{fontenc}

\usepackage[utf8]{inputenc}

\usepackage{microtype}

\usepackage{inconsolata}

\usepackage{graphicx}
\usepackage{url}
\usepackage{booktabs}
\usepackage{multirow}
\usepackage{soul}
\usepackage{amsmath}
\usepackage{xcolor}
\usepackage{promptbox}
\usepackage{subcaption}
\usepackage{tabularx}
\definecolor{gainGreen}{RGB}{0,128,0}
\definecolor{dropRed}{RGB}{200,0,0}
\newcommand{\gain}[1]{\textcolor{gainGreen}{$+#1$}}
\newcommand{\drop}[1]{\textcolor{dropRed}{$-#1$}}

\title{Retrieving Relations, Detecting Fallacies:\\ A RAG Approach to Political Debate Analysis}

\author{
Deborah Dore, Greta Damo, Elena Cabrio \and Serena Villata \\
Université Côte d'Azur, CNRS, INRIA, I3S, France \\
\texttt{\{greta.damo, elena.cabrio\}@univ-cotedazur.fr} \\
\texttt{\{serena.villata, deborah.dore\}@cnrs.fr}
}

\begin{document}
\maketitle
\begin{abstract}

Fallacies are arguments that employ invalid reasoning, making their automatic detection critical in sensitive contexts such as high-stakes political debates, where public opinion is shaped. Spotting a fallacious argument requires contextual knowledge beyond its pure surface text. This entails world knowledge pertaining to the subject matter under discussion, as well as knowledge of the relationships that exist between arguments within the argumentative discourse. Prior work on fallacy analysis has shown that argumentative discourse structure can beneficially improve classification performance. However, such structure is typically encoded only as static classifier features, limiting its flexibility. Building on this intuition while addressing this limitation, we introduce a guided retrieval-augmented methodology for fallacy detection and classification that leverages argumentative relations of support and attack to dynamically steer the extraction of relevant documents. We evaluate our approach on the \texttt{ElecDeb60to20} benchmark across 42 retrieval configurations and 14 models, performing retrieval over a 15GB knowledge base of collected political-related documents. Our approach improves Macro F1 up to 0.864 for fallacy detection and up to 0.725 for classification over non-retrieval baselines. These results show that incorporating external knowledge significantly enhances fallacy detection and classification when retrieval is argumentatively guided.

\end{abstract}

\section{Introduction}
\label{sec:introduction}

Fallacies are argumentative moves that appear valid but violate the principles of correct reasoning \citep{van2001crucial}. 
Though these misleading arguments sound convincing on the surface, their underlying aim is to manipulate public opinion. In this light, recognising fallacies becomes crucial, to enable citizens to engage critically with political information, spot misleading rhetorical strategies, and evaluate candidate claims~\citep{abbas2024fallacy}. A prominent example is \textit{appeal to fear}, where speakers exaggerate hypothetical dangers to trigger an emotional response rather than presenting sound evidence~\cite{walton2010place}. 



Given the potential negative impact of fallacious rhetoric on public discourse, identifying and classifying these arguments remains a key challenge in Argument Mining (AM) \citep{persuasive2017stab, sahai2021breaking, ruiz2023detecting}. 
Recent studies employed argumentative relations as encoded features to improve fallacy detection and classification results~\cite{goffredo2023argument}. However, structural information alone is often insufficient to identify fallacious arguments, as it requires world knowledge, going beyond the surface text of an argument.
To tackle this issue, we introduce a novel method to address the tasks of fallacy detection and classification through \textit{argumentative relation-guided retrieval}. Our approach leverages the underlying argumentative structure, specifically the \textit{support} and \textit{attack} relations between argument components, to dynamically steer external knowledge retrieval. The originality of our approach lies in the fact that, rather than using argumentative structures merely as static input features, we leverage argumentative relations to retrieve highly targeted contextual evidence. Given that we target the political domain, we evaluate our framework on the \texttt{ElecDeb60to20} dataset~\citep{goffredo2023argument}, a comprehensive benchmark spanning televised U.S. presidential debates from 1960 to 2020, annotated with argument components, relations, and fallacies. The main contributions of this paper are: 
\textbf{1) Relation Guided Retrieval for Fallacy Detection and Classification:} we propose a novel method that leverages argumentative relations to steer external knowledge retrieval dynamically, shifting from static structural encodings to context-aware retrieval. We test our approach on the detection of whether an argument is a fallacy or not, and its classification into one of the six most common fallacy types.
\textbf{2) A 15 GB Temporally Grounded Domain Knowledge Base}, tailored to our specific application to U.S. political debates, across six key domains (legislation, entities, geography, history, fallacy taxonomy, and labelled examples). Every record is timestamped, enabling temporally constrained retrieval to improve the focus of historical context. \textbf{3) Extensive Evaluation} of our method across 42 retrieval configurations and 14 models, showing significant gains over state-of-the-art baselines.

\section{Related Work}
\label{sec:related-work}

\paragraph{Fallacy detection and classification.}
In recent years, the NLP community has shown an increasing interest in the task of fallacy detection and classification, and related challenges, such as misinformation and propaganda.

Concerning classification, \citet{habernal2017argotario} released \textit{Argotario}, an open-source software that serves as a gaming platform for educational purposes and as a crowd-sourcing data-acquisition platform to annotate fallacy types in everyday argumentation. Following this study, they released an annotated dataset of fallacious arguments in German and English, and used Support Vector Machine and BiLSTM models to classify fallacy types~\cite{habernal2018adapting}.
A parallel line of research targets propaganda and persuasion techniques, which partly overlap with logical fallacies~\cite{dasanmartino2019fine}. In this work, the authors define an annotation scheme of 18 propaganda techniques and annotate 451 news articles with such labels, and propose a multi-granularity network architecture on top of BERT contextualised embeddings to identify propagandist arguments on different levels of granularity. \citet{jin2022logical} use an architecture based on a simple classifier that includes fallacies' structural information, achieving 0.58 F1 Score on the classification of 13 fallacy classes. \citet{alhindi2022multitask} propose a multitask prompting approach based on the T5 model to identify 28 fallacy types. 
\citet{goffredo2022fallacyclassification} propose a transformer-based neural architecture for detecting and classifying six types of fallacies in U.S. Presidential Election Debates from 1960 to 2016. In subsequent work, \citet{goffredo2023argument} extend this approach by integrating textual, argumentative, and engineered features into a transformer-based architecture, while expanding the original dataset into \texttt{ElecDeb60to20} and consolidating the six-class fallacy taxonomy.

Concerning detection, \citet{vorakitphan2022protect} propose a pipeline for propaganda detection and technique classification, combining semantic and argumentative features, using a BERT-based binary classifier, and a RoBERTa-based model, achieving 0.72 and 0.64 F1 Score, respectively, for 14-class sentence-span classification. Similarly, \citet{sahai2021breaking} address fallacy detection through comment- and token-level classification across eight fallacy types, using Reddit data and contextual information from parent comments. Their finetuned BERT achieves 0.53 Macro F1 on token classification.

With the advancement of Large Language Models (LLMs), more recent work has started to include their knowledge for improving fallacy classification. \citet{sourati2023robust} propose a three-stage framework for fallacy detection, coarse- and fine-grained classification, combining language models with background knowledge and explainable reasoning, highlighting the difficulty of fallacy identification and the need for specialised reasoning across fallacy types. \citet{jeong2025large} use an LLM prompting strategy that enriches input text by incorporating arguments with implicit context and ranks contextual queries by confidence, achieving F1 of 0.57 in zero-shot and 0.45 in finetuned settings across 29 fallacy types. \citet{teo2025large} evaluate a set of LLMs for logical fallacy detection, finding that they perform well on simple fallacies but struggle with more complex, interpretive categories, highlighting the need for improved prompting, finetuning, and contextual data. Finally, \citet{papadopulos2026beyond} combine abstract logical structures with contextual linguistic cues for LLM-based fallacy classification, using data-driven pattern extraction from examples and explanations, improving zero- and one-shot performance. 

In all these cases, however, the added knowledge is hand-crafted or model-generated, never retrieved from an external corpus.

\paragraph{Retrieval Augmented Generation (RAG).}

\citet{lewis2020rag} introduce RAG as a framework for grounding language models in external knowledge retrieved at inference time. While initially developed for knowledge-intensive generation, RAG has since been extended to discriminative tasks, including text classification, where retrieved evidence can provide information beyond the model's parametric knowledge~\citep{gao2023survey}.
Retrieval has also been applied to tasks closely related to fallacy detection, particularly fact checking and argumentation. In fact-checking, benchmarks such as \texttt{FEVER}~\citep{thorne-etal-2018-fever} and \texttt{PubHealth}~\citep{kotonya-toni-2020-explainable-automated} use retrieved evidence to support claim verification. More recently, \citet{dhole-etal-2025-conqret} introduced Retrieval-Augmented Argumentation, showing how external information can be retrieved to support argument quality and groundedness. However, fallacy detection and classification have largely relied on finetuned or few-shot models without explicitly incorporating external knowledge~\citep{cantin-larumbe-chust-vendrell-2025-argumentative}.

In this work, we investigate whether external knowledge can improve fallacy detection and classification in the specific application of political arguments. We adapt RAG to retrieve political knowledge relevant to the argumentative structure of each instance, using information from the relations between argument components to formulate targeted retrieval queries. This allows the model to condition its predictions on task-relevant external evidence rather than relying solely on its parametric knowledge.
\section{Methodology}
\label{sec:methodology}
\begin{figure*}[t]
    \centering
    \includegraphics[width=0.9\linewidth]{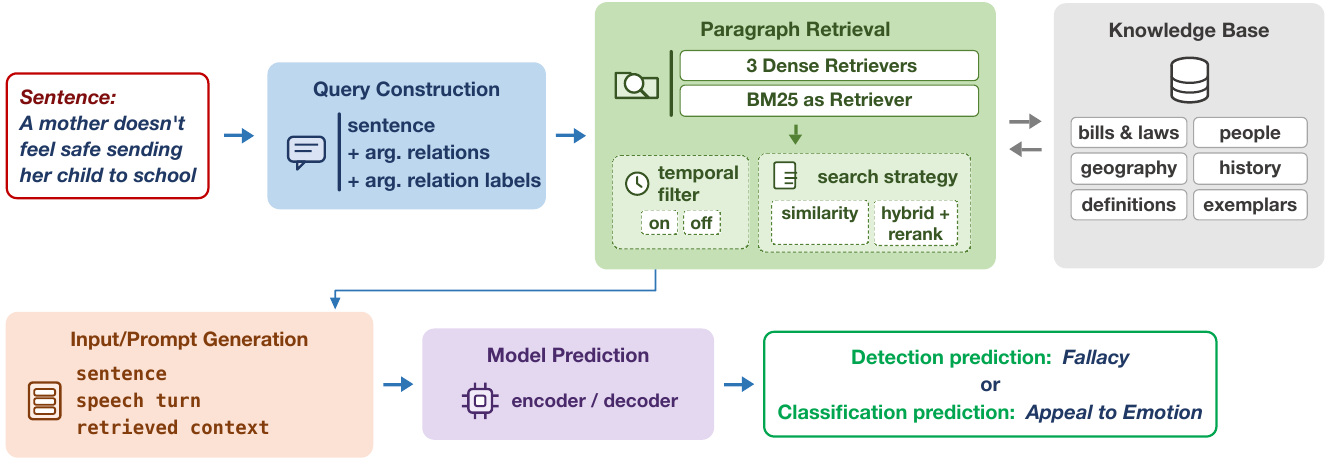}
    \caption{Proposed pipeline for fallacy analysis using external knowledge.}
    \label{fig:pipeline}
\end{figure*}



We consider fallacy detection and fallacy classification as separate tasks to assess how external knowledge contributes to each. In fallacy detection, given a sentence, the goal is to determine whether it contains a fallacious argument. In fallacy classification, given a sentence known to be fallacious, the goal is to assign it to exactly one of six fallacy types: Ad Hominem, Appeal to Authority, Appeal to Emotion, Slippery Slope, False Cause, or Slogan. To address these tasks, we propose a four-stage pipeline comprising: query generation, retrieval, input generation and model prediction (Figure \ref{fig:pipeline}). %

The \textbf{query generation} stage progressively incorporates information from the argumentative structure, resulting in three query variants:

\textit{(i) sentence-only query:} the baseline query consists of the target sentence whose fallaciousness or fallacy type is to be predicted. When no relation is annotated, only this query strategy applies;

\textit{(ii) sentence+relations query:} since identifying a fallacy often depends on its argumentative context, we exploit the relation layer. For each component, we append its related components to the target sentence, thereby providing the retriever with a broader view of the local argument;

\textit{(iii) sentence+relations+labels query:} we further incorporate the labels of the argumentative relations (support or attack). Rather than representing these labels as bare tokens, we verbalise each relation as an explicit clause of the form \texttt{<related> supports/attacks <target>}, thereby encoding the direction of the argumentative relation directly. 

During the \textbf{retrieval} stage, the retriever uses the generated query to retrieve relevant external evidence from the knowledge base (KB). The retrieved evidence is then combined with the local debate context to form a \textbf{single model input}, which is provided to the model for \textbf{prediction}. The proposed pipeline is not tied to a specific domain and can be extended to other settings where the relevant argumentative relations are available.

To isolate the contribution of argumentative relations from that of retrieval, we tested an additional setting in which the relations are provided directly in the model's input, with no retrieval (see Appendix \ref{sec:input-relation-enriched}).

\section{Resources for Fallacy Analysis of Political Debates}

\subsection{Dataset}
 To evaluate the proposed method, we choose the \texttt{ElecDeb60to20} dataset~\footnote{\url{https://github.com/MARIANNE-INRIA/ElecDeb60to20}}, 
%
to the best of our knowledge, the only available resource for political debates annotated with argumentative components (15,093 claims and 13,623 premises), argumentative component relations (20,805 support and 3,701 attack relations), and argumentative fallacies. 
The dataset contains the annotations for the six most common types of fallacies in political debate: \textbf{Ad Hominem} (an excessive attack on the arguer rather than their position, 200 examples), \textbf{Appeal to Emotion} (loading the argument with emotional language to exploit the audience, 924 examples), \textbf{Appeal to Authority} (citing an authority, a non-expert, or majority acceptance in place of relevant evidence, 290 examples), \textbf{Slippery Slope} (claiming an unlikely, exaggerated outcome will follow an act, 72 examples), \textbf{False Cause} (mistaking correlation for causation, 105 examples), and \textbf{Slogans} (a brief, striking phrase meant to excite the audience, 42 examples).
\subsection{Knowledge Base}
\label{sec:KB}
\paragraph{Data collection.} To ground retrieval, we first assembled a collection of documents relevant to the political scenario to create the KB.

We analysed the debate transcripts of the above-mentioned dataset along two axes. First, we run the spaCy \texttt{en\_core\_web\_trf} named-entity recogniser over every debate and group the recognised mentions into people, organisations, events and places; after normalisation and de-duplication this yields 1{,}081 people, 974 organisations, 80 events and 595 places. Second, we perform \textbf{topic extraction}: we segment each debate into non-overlapping 1{,}024-token windows, prompt Qwen2.5-7B-Instruct~\citep{qwen2025} to return the main topics of each window in a fixed \texttt{<TOPICS: t1, t2, ...>} format, parse the generations with a regular expression, and lower-case and merge them across windows. This gives 2{,}057 unique topics and 3{,}115 distinct (topic, debate-year) pairs. 

Guided by this analysis we assemble six collections of documents~\footnote{All material was crawled between May and June 2026.}, each targeting a distinct kind of grounding: \textbf{Legislation} comes from the U.S. Government Information website~\footnote{\url{https://www.govinfo.gov}}, whose search API we query by (topic, year), expanding each topic over a five-year window $[y-4, y]$ and keeping the 15 highest-scoring granules per pair, so as to retrieve the bills and public laws a debate of year $y$ could plausibly be about. We complement these with the 2025 United States Government Manual and the Universal Declaration of Human Rights.

The \textbf{people} collection grounds the debates in the actors they name, combining official biographies from the Biographical Directory of the United States Congress~\footnote{\url{https://bioguide.congress.gov}} with the Wikipedia~\footnote{\url{https://en.wikipedia.org}} pages of the individuals mentioned in the debates, expanded with the first-level pages linked from each article.

The \textbf{geography} collection draws on official U.S. Census Bureau state and territory profiles~\footnote{\url{https://data.census.gov}}, including demographic, economic and social statistics, together with the Wikipedia pages of the places named in the debates and their linked pages.

For temporal grounding, the \textbf{history} collection covers U.S. history from 1900 to 2026 with one record per year, extracted from the \textit{Events} section of the corresponding \textit{``\{year\} in the United States''} Wikipedia page and complemented by the era-overview pages of \textit{History of the United States} and their first-level links. Every record is annotated with its year, so each chunk keeps its temporal context after splitting.

The \textbf{definitions} collection supplies the theory of fallacy itself: scholarly articles published from 2015 onwards on fallacies, argumentation, rhetoric and political discourse, retrieved from the OpenAlex academic index~\footnote{\url{https://api.openalex.org}}, restricted to \texttt{type:article} and filtered by lexical relevance over title and abstract, together with the Wikipedia \textit{Fallacy} page and its relevant linked articles.

Finally, the \textbf{exemplars} collection holds labelled fallacy instances drawn from \texttt{LOGIC}~\citep{jin2022logical} and \texttt{MAFALDA}~\citep{helwe2024mafalda}, mapped onto our six-type taxonomy. Because these records carry gold fallacy labels, we verify that no exemplar text overlaps an \texttt{ElecDeb60to20} instance and remove any that does, so that retrieval cannot leak labels into the evaluation. Table~\ref{tab:KB-collection-table} reports the resulting statistics. 

\begin{table}[h]
\centering
\resizebox{\columnwidth}{!}{%
\begin{tabular}{@{}lccc@{}}
\toprule
\textbf{Collection} & \textbf{Size} & \textbf{\# Documents} & \textbf{\# Paragraphs} \\ \midrule
Bills       & 14 GB  &   178{,}383 &     6{,}668{,}178 \\
Geography   & 516 MB &   125{,}891 &         125{,}891 \\
People      & 297 MB &    72{,}950 &          72{,}950 \\
History     &  89 MB &    16{,}121 &          16{,}121 \\
Exemplars   &   5 MB &     1{,}273 &           1{,}284 \\
Definitions & 1.3 MB &          65 &                65 \\
\midrule
\textbf{Total} & \textbf{15 GB} & \textbf{394{,}683} & \textbf{6{,}884{,}489} \\
\bottomrule
\end{tabular}%
}
\caption{Statistics of the collected KB documents.}
\label{tab:KB-collection-table}
\end{table}

The collection is dominated in volume by \textbf{bills}, which account for the vast majority of both its size and its paragraphs, a direct consequence of the length and density of legislative text. \textbf{Geography} and \textbf{people} contribute the next largest document counts, each with a single paragraph per record. \textbf{History}, \textbf{exemplars} and \textbf{definitions} are comparatively tiny, but play a targeted role, supplying temporal grounding and fallacy-specific knowledge respectively.

\paragraph{Indexing.}
We index the KB with LangChain~\footnote{\url{https://www.langchain.com}}. Every document is normalised into a JSON record with a unique identifier (\texttt{id}), its \texttt{text}, and the metadata used later to constrain retrieval: \texttt{source}, \texttt{URL}, \texttt{year} and \texttt{category} (the collection).

Long documents are split into chunks of at most 1{,}000 tokens with a 10\% overlap. We chunk by tokens rather than by characters and clamp the chunk size to the input window of the embedding model in use, so that no passage is silently truncated at encoding time. Chunks are embedded and indexed in Chroma~\footnote{\url{https://www.trychroma.com}}, one store per embedding model, with the document set held fixed so that retrievers can be compared in isolation.

\paragraph{Retrievers.}
We compare three dense encoders against a lexical baseline. \textbf{Sentence-BERT (SBERT)}~\citep{reimers2019sbert} encodes queries and passages into embeddings optimised for semantic similarity. \textbf{BAAI General Embedding (BGE)}~\citep{xiao2024cpackpackedresourcesgeneral} is trained on a large multi-task retrieval corpus and provides a stronger general-purpose semantic space. As a third dense retriever, we decided to pretrain a sentence transformer tailored to our political domain: \textbf{Sentence-RooseBERT}~\footnote{\url{huggingface.co/MARIANNE-INRIA/Sentence-RooseBERT}}, trained with the corpus and hyperparameters defined by \citet{dore2025roosebert}. \textbf{BM25}~\citep{robertson2009probabilistic} ranks passages by exact term overlap.

\paragraph{Search strategies.}
At inference, we retrieve the top $k{=}5$ chunks per query (as preliminary experiments showed this is the optimal number). In the \textbf{similarity} setting these come directly from the dense index by cosine similarity. In the \textbf{hybrid} setting we fuse the dense and BM25 rankings with Reciprocal Rank Fusion ($k_{\text{RRF}}{=}60$) over a pool of 30 candidates per leg, then rerank the fused pool with a \texttt{ms-marco-MiniLM-L-6-v2} cross-encoder before keeping the top 5. Each of the three dense retrievers is evaluated with the two search strategies while BM25 is evaluated with its lexical search strategy. Each retriever is then evaluated with the three query formulations.

Here we introduce a \textit{temporal filtering} constraint: we test whether restricting the KB to the years preceding each debate helps retrieve more focused, contextually appropriate knowledge.

\section{Experimental Setting}
\label{sec:experimental-setting}

\subsection{Tasks}
For the task of \textbf{fallacy detection}, we sentence-split the speech turns with spaCy and propagate the span-level annotations down to the sentence level, labelling a sentence \textit{Fallacy} if it overlaps at least one annotated fallacy span and \textit{NoFallacy} otherwise. The resulting distribution is strongly skewed towards the negative class, so we downsample \textit{NoFallacy} to a 1:1 ratio and re-split the balanced pool 80/10/10, stratified by label, yielding 2{,}612 training, 327 validation and 327 test sentences.

In \textbf{fallacy classification}, the task is restricted to the fallacious portion of the corpus (1{,}306 training, 164 validation and 163 test sentences) and remains sharply imbalanced: \textit{Appeal to Emotion} alone accounts for more than half of the training instances (739), while \textit{Slogans} accounts for 34.

\subsection{Models}
\label{subsec:models}
On both tasks, we evaluate \textbf{encoder-only} models finetuned as sentence classifiers, and \textbf{decoder-only} LLMs that output the label as text. 
For encoder-only models, the input is formatted as: \texttt{sentence [SEP] speech turn context}. In particular, we evaluated RoBERTa \citep{liu2019roberta}, DeBERTa-v3~\citep{he2023deberta}, ModernBERT~\citep{warner2025modernbert}, NeoBERT~\citep{breton2025neobert} and Longformer~\citep{beltagy2020longformer}, alongside two specialised encoders: LegalBERT~\citep{zheng2021legalbert}, chosen because a large share of our KB is legislative text, and RooseBERT~\citep{dore2025roosebert}, pretrained on political and parliamentary debate transcripts. 
RooseBERT is evaluated in all variants (\texttt{scr}/\texttt{cont} $\times$ cased/uncased). Encoders are finetuned over a grid of batch size $\{8, 16\}$, learning rate $\{2, 3, 5\}\times e^{-5}$ and $\{3, 5\}$ epochs, with a warmup ratio of $0.06$.

For decoder-only models, the prompt includes: a system message with the task and taxonomy (for classification), the target sentence, the debate turn context, the retrieved context, and an instruction to output a single label. We evaluated \texttt{Llama-3.1-8B-Instruct}~\citep{grattafiori2024llama3herdmodels}, \texttt{Mistral-7B-Instruct-v0.3}~\citep{jiang2023mistral7b}, \texttt{Qwen3.5-9B-Base}~\citep{qwen2025} and \texttt{gpt-oss-20b}~\citep{openai2025gptoss}. Each is run in three settings: \texttt{zero-shot}, \texttt{few-shot} and \texttt{finetune}.
In \texttt{zero-shot} the model sees only the task instruction and the input. In \texttt{few-shot} we prepend a class-balanced set of labelled examples drawn from the training split, one per class. 
In the last setting, we perform parameter-efficient supervised fine-tuning with LoRA~\citep{hu2022lora} ($r{=}16$, $\alpha{=}32$, dropout $0.05$) over learning rate $\{1, 3, 5\}\times e^{-4}$ and $\{1, 2, 3\}$ epochs at an effective batch size of $8$. 

\subsection{Implementation}
\label{subsec:protocol}
Combining retrievers, search strategies, query formulations and the temporal constraints yields 42 configurations. Adding the no-retrieval baseline gives \textbf{43 configurations per task}.
Evaluating 43 configurations $\times$ $14$ models $\times$ a hyperparameter grid $\times$ five seeds is not computationally feasible, so we proceed as follows:
\textit{(i)} Every model is trained on the no-retrieval baseline across the full hyperparameter grid. The best configuration per model is then re-run with five seeds.
\textit{(ii)} The three strongest baseline models per task are then run over all remaining 42 retrieval configurations, identifying the best retrieval configuration for that task. 
\textit{(iii)} Every model is finally re-run on the best setup, sweeping over hyperparameters and then five seeds.

All experiments use Hugging Face with Python 3.10. Encoders were trained on one H100 GPU and the LLMs on two A100 GPUs. To pretrain Sentence-RooseBERT we used eight A100 GPUs and the same hyperparameters defined by \citet{dore2025roosebert}. Code is available at \url{https://github.com/deborahdore/FallacyRAG}.

\subsection{Metrics}
\label{subsec:metrics}
Both tasks are evaluated with Accuracy and macro-averaged Precision, Recall and F1. We take \textbf{Macro F1} as the primary metric, since it weights every class equally. All scores are averaged over five runs (seeds $\{3, 14, 42, 55, 89\}$), and we report mean and standard deviation. Differences between the best baseline and retrieval model are tested with a two-sided paired $t$-test over the five paired per-seed Macro F1 values, reported with Cohen's $d$.

\begin{table*}[ht]
\centering
\small
\setlength{\tabcolsep}{4pt}
\begin{tabular}{llcccccc}
\toprule
& & \multicolumn{3}{c}{\textbf{Detection}} & \multicolumn{3}{c}{\textbf{Classification}} \\
\cmidrule(lr){3-5} \cmidrule(lr){6-8}
\textbf{Model} & \textbf{Setting} & Baseline & RAG & $\Delta$ & Baseline & RAG & $\Delta$ \\
\midrule
RoBERTa-base & finetune & $74.7 \pm 0.8$ & $80.9 \pm 1.5$ & \gain{6.2} & $34.5 \pm 25.0$ & $68.0 \pm 2.3$ & \gain{33.5} \\
DeBERTa-v3-base & finetune & $74.2 \pm 6.1$ & $77.5 \pm 2.2$ & \gain{3.3} & $43.5 \pm 13.0$ & $45.5 \pm 19.7$ & \gain{2.0} \\
ModernBERT-base & finetune & $76.4 \pm 1.7$ & $\mathbf{86.4 \pm 1.2}$ & \gain{10.0} & $57.7 \pm 2.1$ & $\mathbf{72.5 \pm 4.0}$ & \gain{14.8} \\
NeoBERT & finetune & $57.2 \pm 21.9$ & $50.8 \pm 24.0$ & \drop{6.4} & $24.9 \pm 6.8$ & $10.8 \pm 9.8$ & \drop{14.1} \\
LegalBERT & finetune & $71.7 \pm 1.5$ & $77.4 \pm 1.7$ & \gain{5.7} & $53.9 \pm 23.7$ & $60.8 \pm 3.7$ & \gain{6.9} \\
Longformer-base-4096 & finetune & $74.5 \pm 2.0$ & $78.7 \pm 1.9$ & \gain{4.2} & $63.3 \pm 2.0$ & $58.2 \pm 4.8$ & \drop{5.1} \\
RooseBERT-scr-cased & finetune & $\mathbf{77.2 \pm 1.2}$ & $78.8 \pm 1.1$ & \gain{1.6} & $63.0 \pm 2.4$ & $69.1 \pm 2.2$ & \gain{6.1} \\
RooseBERT-scr-uncased & finetune & $69.5 \pm 0.9$ & $74.3 \pm 1.8$ & \gain{4.8} & $40.1 \pm 4.6$ & $27.0 \pm 2.7$ & \drop{13.1} \\
RooseBERT-cont-cased & finetune & $74.7 \pm 2.3$ & $77.9 \pm 1.1$ & \gain{3.2} & $\mathbf{65.3 \pm 4.3}$ & $68.8 \pm 1.5$ & \gain{3.5} \\
RooseBERT-cont-uncased & finetune & $74.2 \pm 2.5$ & $77.0 \pm 2.7$ & \gain{2.8} & $44.2 \pm 12.4$ & $54.6 \pm 24.4$ & \gain{10.4} \\
\midrule
\multirow{3}{*}{Llama-3.1-8B-Instruct}
 & zero-shot & $56.4 \pm 0.0$ & $54.3 \pm 0.0$ & \drop{2.1} & $36.4 \pm 0.0$ & $22.7 \pm 0.0$ & \drop{13.7} \\
 & few-shot & $59.0 \pm 0.0$ & $33.4 \pm 0.0$ & \drop{25.6} & $3.6 \pm 0.0$ & $0.0 \pm 0.0$ & \drop{3.6} \\
 & finetune & $70.7 \pm 1.2$ & $63.0 \pm 2.3$ & \drop{7.7} & $53.3 \pm 1.9$ & $20.2 \pm 1.7$ & \drop{33.1} \\
\addlinespace
\multirow{3}{*}{Mistral-7B-Instruct-v0.3}
 & zero-shot & $56.1 \pm 0.0$ & $41.1 \pm 0.0$ & \drop{15.0} & $31.5 \pm 0.0$ & $19.7 \pm 0.0$ & \drop{11.8} \\
 & few-shot & $59.1 \pm 0.0$ & $33.4 \pm 0.0$ & \drop{25.7} & $1.6 \pm 0.0$ & $3.6 \pm 0.0$ & \gain{2.0} \\
 & finetune & $57.5 \pm 8.3$ & $47.0 \pm 3.3$ & \drop{10.5} & $34.3 \pm 12.9$ & $19.1 \pm 4.3$ & \drop{15.2} \\
\addlinespace
\multirow{3}{*}{Qwen3.5-9B-Base}
 & zero-shot & $33.4 \pm 0.0$ & $33.4 \pm 0.0$ & $0.0$ & $1.6 \pm 0.0$ & $1.2 \pm 0.0$ & \drop{0.4} \\
 & few-shot & $33.4 \pm 0.0$ & $33.4 \pm 0.0$ & $0.0$ & $0.0 \pm 0.0$ & $0.0 \pm 0.0$ & $0.0$ \\
 & finetune & $33.5 \pm 0.2$ & $33.4 \pm 0.0$ & \drop{0.1} & $1.3 \pm 1.9$ & $0.0 \pm 0.0$ & \drop{1.3} \\
\addlinespace
\multirow{3}{*}{gpt-oss-20b}
 & zero-shot & $33.4 \pm 0.0$ & $33.4 \pm 0.0$ & $0.0$ & $0.0 \pm 0.0$ & $0.0 \pm 0.0$ & $0.0$ \\
 & few-shot & $33.4 \pm 0.0$ & $33.4 \pm 0.0$ & $0.0$ & $0.0 \pm 0.0$ & $0.0 \pm 0.0$ & $0.0$ \\
 & finetune & $63.8 \pm 2.5$ & $54.0 \pm 3.3$ & \drop{9.8} & $34.3 \pm 7.4$ & $13.4 \pm 3.9$ & \drop{20.9} \\
\bottomrule
\end{tabular}
\caption{Macro F1 (mean $\pm$ std, 5 seeds) results. RAG uses a Sentence-RooseBERT retriever with hybrid search (temporal context for detection, none for classification). $\Delta =$ RAG $-$ Baseline. Best model per column in \textbf{bold}.}
\label{tab:results-fallacy-combined}
\end{table*}
\section{Results}
\label{sec:results}
Table \ref{tab:results-fallacy-combined} reports the results of the two tasks.

\paragraph{Fallacy Detection.}
\label{subsec:res-detection}
Without retrieval, the strongest model is \texttt{RooseBERT-scr-cased} ($77.2$ Macro F1), followed by \texttt{ModernBERT} ($76.4$) and \texttt{RooseBERT-cont-cased} ($74.7$).
The best of the 42 configurations combines temporal filtering, \texttt{Sentence-RooseBERT} as retriever, hybrid search with cross-encoder reranking, and \textit{sentence+relations+labels} as the query. Under it, \texttt{ModernBERT} reaches $\mathbf{86.4}$ macro F1, $+10.0$ over its own baseline and $+9.2$ over the best baseline model. A two-sided paired $t$-test over the five seeds confirms the difference is significant ($t = 19.17$, $p = 4\times10^{-5}$, Cohen's $d = 8.57$).

Retrieval improves every encoder we test, from $+1.6$ (\texttt{RooseBERT-scr-cased}) to $+10.0$ (\texttt{ModernBERT}), with the sole exception of \texttt{NeoBERT}, which fails to converge reliably in either setting. It also stabilises training: the standard deviation of \texttt{DeBERTa-v3} falls from $6.1$ to $2.2$. 

Decoder-only models behave in the opposite way. \texttt{Llama}, \texttt{GPT-oss} and \texttt{Mistral} each lose between $2.1$ to $25.7$ Macro F1 once retrieved context is added, despite competitive baselines.

\paragraph{Fallacy Classification.}
\label{subsec:res-classification}
The strongest baseline model is \texttt{RooseBERT-cont-cased} ($65.3$ Macro F1), followed by \texttt{Longformer} ($63.3$) and \texttt{RooseBERT-scr-cased} ($63.0$). Six of the fourteen models fall below $35.0$, and baseline variance is markedly higher than in detection: five encoders exceed a standard deviation of 5 across seeds (\texttt{RoBERTa}, \texttt{DeBERTa-v3}, \texttt{NeoBERT}, \texttt{LegalBERT}, \texttt{RooseBERT-cont-uncased}).

The best configuration again uses \texttt{Sentence-RooseBERT} with hybrid search, cross-encoder reranking and the \textit{sentence+relations+labels} query, while temporal filtering is not applied. Under it, \texttt{ModernBERT} reaches $\mathbf{72.5}$ Macro F1, $+14.8$ over its own baseline and $+7.2$ over the best baseline model ($t(4)=4.62$, $p=0.010$, $d_z=2.07$). Retrieval therefore helps on classification as well as on detection, though the margin is narrower.

Part of the gain comes from stabilising weak models: \texttt{RoBERTa} improves by $+33.5$ while its standard deviation falls, and \texttt{LegalBERT} gains $6.9$ with standard deviation down from $23.7$ to $3.7$. Retrieved context appears to supply discriminative signal that the target sentence alone does not carry for fine-grained types.


Decoder-only models degrade sharply. \texttt{Llama} loses $33.1$ and \texttt{Mistral} $15.2$ Macro F1, both in finetuning settings. Inspection of per-class scores shows this is not a single-class collapse but uniformly low performance across all six fallacy types. The retrieved passages are long, domain-heavy, and unsupervised at inference time. A finetuned encoder learns which parts of them matter, a prompted model is simply crowded out.

\paragraph{Analysis of the retrieved knowledge.}
We analyse which KB collections are retrieved for each task, reporting the share of training arguments that retrieve at least one document from each collection (an argument can retrieve from several, so shares do not sum to 100\%).
\textit{Bills} are the dominant source for both tasks, but far more so for classification (93\% vs.\ 76\%).
Detection instead draws on a broader range of contextual knowledge: \textit{people} (32\% vs.\ 10\%), \textit{geography} (26\% vs.\ 4\%), and \textit{history} (13\% vs.\ 5\%).
\textit{Examples} are retrieved less often (8\% and 3\%), and \textit{definitions} and \textit{papers} are almost never retrieved ($<$1\% for both tasks).
This suggests that deciding \emph{whether} an argument is fallacious benefits from grounding its factual context, whereas classification concentrates on the policy content under debate.
The near absence of \textit{definitions} and \textit{papers} further indicates that retrieval is driven by topical similarity rather than by conceptual knowledge about fallacies.

\paragraph{Error Analysis.}
We conducted an error analysis comparing the best baseline against the best retrieval model on both tasks (\texttt{RooseBERT-scr-cased} vs \texttt{ModernBERT} for detection and \texttt{RooseBERT-cont-cased} vs \texttt{ModernBERT} for classification). Detection and classification benefit from retrieval differently. Temporal filtering helps detection but hurts classification, suggesting that deciding \emph{whether} an argument is fallacious relies on contemporaneous factual context, while classifying \emph{which} fallacy depends on rhetorical form, largely era-independent. Restricting the knowledge base thus removes useful exemplars for classification without providing relevant context.

For the detection, our findings show that \texttt{RooseBERT} is biased towards the positive class: 14.98\% of the test set consists of non-fallacious arguments predicted as fallacious, almost twice the rate of \texttt{ModernBERT} (7.65\%).
Retrieval reduces both false positives and false negatives by half, 8.87\% and 3.67\% respectively, for an overall error rate of 12.54\% against the 22.63\% of the baseline.
The gain for the retrieval is largest on the \textit{NoFallacy} class (recall from 0.7 to 0.82) which also raises precision on the \textit{Fallacy} class from 0.74 to 0.84.
This suggests that retrieved context is especially useful for disambiguating rhetorically marked, non-fallacious arguments.

For the classification, both models over-predict \textit{Appeal to Emotion}, which makes up 56.4\% of gold labels and accounts for 53\% of baseline errors and 55\% of errors with retrieval. Both models also confuse \textit{Appeal to Authority} with \textit{Appeal to Emotion} at similar rates, suggesting this is not a knowledge gap: speakers often cite authorities emotionally, so the two classes look alike on the surface, and added context does not separate them.
Retrieval helps classes defined by form rather than content: \textit{Slogans} recall rises from 0.25 to 0.75, and \textit{Slippery Slope} precision improves from 0.40 to 0.57. This comes at a cost: \textit{Ad Hominem} precision drops from 0.86 to 0.69 and \textit{False Cause} recall from 0.80 to 0.50, so accuracy falls by 1.2 points even as Macro F1 gains 3.9 points. Retrieval, then, does not reduce errors uniformly: it redistributes them.

This pattern reflects the nature of our knowledge base, dominated by legislation, history, and general knowledge. Such sources effectively disambiguate fallacious from non-fallacious arguments but provide fewer useful exemplars for fine-grained type classification, which benefits more from examples.

Across both tasks, \texttt{Sentence-RooseBERT} consistently outperforms other retrievers, likely because domain-specific pretraining on parliamentary data aligns better with debate discourse than general-purpose embeddings. 

\section{Ablation Study}
\label{sec:ablation}
The proposed pipeline combines four components: a structure-informed query, a dense retriever, hybrid fusion with cross-encoder reranking, and temporal filtering of the KB. To measure what each contributes we change one component at a time and leave the rest untouched (see Table~\ref{tab:ablation-elecdeb}).

\begin{table}[ht]
\centering
\footnotesize
\setlength{\tabcolsep}{4pt}
\begin{tabularx}{\columnwidth}{@{}>{\raggedright\arraybackslash}Xrr@{}}
\toprule
\textbf{Configuration} & \textbf{Detection} & \textbf{Classification} \\
\midrule
Full pipeline & $\mathbf{80.2}$ & $\mathbf{71.8}$ \\
\midrule
\multicolumn{3}{@{}l}{\textit{Retrieval query}} \\
\hspace{0.8em}$\rightarrow$ \texttt{sentence+relations} & $78.3$ & $61.0$ \\
\hspace{0.8em}$\rightarrow$ target \texttt{sentence} only & $73.0$ & $66.4$ \\
\midrule
\multicolumn{3}{@{}l}{\textit{Retriever}} \\
\hspace{0.8em}$\rightarrow$ BGE & $77.8$ & $57.9$ \\
\hspace{0.8em}$\rightarrow$ SBERT & $78.6$ & $60.1$ \\
\hspace{0.8em}$\rightarrow$ BM25 (sparse only) & $75.7$ & $60.9$ \\
\hspace{0.8em}\hspace{0.8em}$\hookrightarrow$ + sentence query & $72.7$ & $59.1$ \\
\midrule
\multicolumn{3}{@{}l}{\textit{Fusion and reranking}} \\
\hspace{0.8em}$\rightarrow$ dense similarity only & $76.3$ & $62.1$ \\
\midrule
\multicolumn{3}{@{}l}{\textit{Temporal filtering}} \\
\hspace{0.8em}$\rightarrow$ \texttt{off} for detection, \texttt{on} for classification & $77.5$ & $49.8$ \\
\midrule
\multicolumn{3}{@{}l}{\textit{Retrieval}} \\
\hspace{0.8em}$\rightarrow$ no retrieval & $75.4$ & $65.3$ \\
\bottomrule
\end{tabularx}
\caption{Results of the ablation study (Macro F1) with hyperparameters frozen. Single seed (14).} 
\label{tab:ablation-elecdeb}
\end{table}

Each ablation is anchored on the strongest model \textit{without} retrieval: \texttt{RooseBERT-scr-cased} for detection, \texttt{RooseBERT-cont-cased} for classification, with frozen hyperparameters: batch size of 8, learning rate of 5e-5, trained for 5 epochs.

On detection the full pipeline reaches $80.2$ Macro F1 against $75.4$ without retrieval ($+4.8$). Replacing the relation query with only the target sentence is the most damaging change ($-7.2$), and pushes the system $2.4$ points \emph{below} the no-RAG baseline: retrieving only on the sentence is worse than not retrieving at all. The effect is not a property of the embedding model. BM25 as retriever with a structured query reaches $75.7$, the same as with the no-RAG baseline, but BM25 retrieval with the target sentence falls to $72.7$. Exchanging one dense retriever for another, by contrast, costs only $1.6$--$2.4$ points, and every dense variant stays above the baseline. Dense retrieval is thus necessary but not sufficient: it is worth $4.5$ points over BM25 at a matched query, roughly half of what the query itself is worth, and no choice of embedding model recovers the gain once the query loses its argumentative structure. Reranking sits in the same range ($-3.9$), marginally above the baseline.

On classification the full pipeline reaches $71.8$ against $65.3$ ($+6.5$) for the baseline. Swapping the dense retriever is the most expensive change to the retrieval stack ($-13.9$ for BGE, $-11.7$ for SBERT): on this task no alternative retriever, sparse or dense, reproduces what Sentence-RooseBERT retrieves. The ordering of the query modes also inverts with respect to detection: \texttt{sentence+relations} ($61.0$) is worse than \texttt{sentence-only} ($66.4$), which is the only configuration to stay above the baseline. We read this as evidence that the gain on classification arises from a specific retriever-query pairing rather than from retrieval quality in general.

\section{Additional Experiments}
\label{additional-experiments}
To assess whether our approach generalises beyond \texttt{ElecDeb60to20}, we looked for a dataset annotated with fallacies whose domain overlaps with our knowledge base.
We selected \texttt{PROPAGANDA}~\cite{dasanmartino2019fine}, a corpus of news articles annotated at the fragment level with eighteen propaganda techniques. We consider the sentence-level task, i.e., the binary detection of propagandistic sentences, for which the authors report a Macro F1 of 60.98. \texttt{PROPAGANDA} carries no relation layer, but every sentence is linked to the news article it comes from, which gives us the candidate components needed to recover relations automatically.

We thus trained a relation classifier on the most similar relation-annotated corpus we could find and applied it to \texttt{PROPAGANDA}. That corpus is \texttt{LIARArg}~\cite{wang2025automated}, 2,832 news claims, annotated with argument components (claim, premise) and fine-grained relations (support, attack, partial support, partial attack). Following the original work, we merge the partial labels into \textit{support} and \textit{attack}, yielding a three-way \textit{support}/\textit{attack}/\textit{no\_relation} task. We trained all the models of Section~\ref{subsec:models} on \texttt{LIARArg} and selected the one with the highest Macro F1 averaged over 5 seeds (see Table~\ref{tab:liar-results}, Appendix \ref{appendix:liar-resuls}).

\texttt{Llama}, the best relation classifier (see Table \ref{tab:liar-results}), was used to label \texttt{PROPAGANDA}: each propaganda-annotated sentence was paired with every other sentence of its own article, and the predicted \textit{support} and \textit{attack} relations were then used to build the retrieval queries as in our main experiments. 

We tested the configuration that performed best on \texttt{ElecDeb60to20} (dense retriever, hybrid search, reranking, and the \textit{sentence+relations+labels} query) and re-tuned only the retriever, running all three dense models on this dataset: here \texttt{SBERT} turned out to be the best choice. The same protocol described in Section~\ref{subsec:protocol} was followed.
\begin{table}[h]
\resizebox{\columnwidth}{!}{%
\begin{tabular}{llccc}
\toprule
\textbf{Model} & \textbf{Setting} & Baseline & RAG & $\Delta$ \\
\midrule
RoBERTa-base             & finetune & 73.5 $\pm$ 0.6  & 85.0 $\pm$ 0.8 & \gain{11.5} \\
DeBERTa-v3-base          & finetune & 74.5 $\pm$ 0.5  & 42.7 $\pm$ 0.0 & \drop{31.8} \\
ModernBERT-base          & finetune & 73.3 $\pm$ 0.5  & 81.3 $\pm$ 0.8 & \gain{8.0}  \\
NeoBERT                  & finetune & 53.5 $\pm$ 14.7 & 42.7 $\pm$ 0.0 & \drop{10.8} \\
LegalBERT                & finetune & 73.0 $\pm$ 0.4  & 84.9 $\pm$ 0.3 & \gain{11.9} \\
Longformer-base-4096     & finetune & 73.7 $\pm$ 0.4  & \textbf{85.5 $\pm$ 1.4} & \gain{11.8} \\
RooseBERT-scr-cased      & finetune & 72.7 $\pm$ 0.4  & 83.6 $\pm$ 0.5 & \gain{10.9} \\
RooseBERT-scr-uncased    & finetune & 74.1 $\pm$ 0.4  & 83.7 $\pm$ 0.8 & \gain{9.6}  \\
RooseBERT-cont-cased     & finetune & 73.4 $\pm$ 0.6  & 84.2 $\pm$ 0.3 & \gain{10.8} \\
RooseBERT-cont-uncased   & finetune & 73.2 $\pm$ 0.8  & 83.1 $\pm$ 2.7 & \gain{9.9}  \\
\midrule
\multirow{3}{*}{Llama-3.1-8B-Instruct}
 & zero-shot & 70.0 $\pm$ 0.0 & 57.8 $\pm$ 0.0 & \drop{12.2} \\
 & few-shot  & 63.8 $\pm$ 0.0 & 42.7 $\pm$ 0.0 & \drop{21.1} \\
 & finetune & 74.8 $\pm$ 0.3 & 47.6 $\pm$ 2.3 & \drop{27.2} \\
\addlinespace
\multirow{3}{*}{Mistral-7B-Instruct-v0.3}
 & zero-shot & 59.7 $\pm$ 0.0 & 37.7 $\pm$ 0.0 & \drop{22.0} \\
 & few-shot  & 62.0 $\pm$ 0.0 & 42.7 $\pm$ 0.0 & \drop{19.3} \\
 & finetune & \textbf{74.9 $\pm$ 0.9} & 44.1 $\pm$ 2.4 & \drop{30.8} \\
\addlinespace
\multirow{3}{*}{Qwen3.5-9B-Base}
 & zero-shot & 42.7 $\pm$ 0.0  & 42.8 $\pm$ 0.0 & \gain{0.1} \\
 & few-shot  & 42.7 $\pm$ 0.0  & 42.7 $\pm$ 0.0 & 0.0        \\
 & finetune & 50.1 $\pm$ 14.3 & 42.8 $\pm$ 0.1 & \drop{7.3} \\
\addlinespace
\multirow{3}{*}{gpt-oss-20b}
 & zero-shot & 42.7 $\pm$ 0.0 & 42.7 $\pm$ 0.0 & 0.0 \\
 & few-shot  & 42.7 $\pm$ 0.0 & 42.7 $\pm$ 0.0 & 0.0 \\
 & finetune & 73.5 $\pm$ 1.1 & 42.8 $\pm$ 0.1 & \drop{30.7} \\
\bottomrule
\end{tabular}%
}
\caption{Macro F1 (mean $\pm$ std, 5 seeds) results. $\Delta =$ RAG $-$ Baseline. Best model per column in \textbf{bold}.}
\label{tab:propaganda-results}
\end{table} 

In Table~\ref{tab:propaganda-results}, we show that our approach improves over the baseline by 10.6 points of Macro F1 (24.5 over the score reported by the original authors), confirming that the benefit of relation-aware retrieval is not specific to \texttt{ElecDeb60to20}. The difference between the best baseline model (\texttt{Mistral} with 74.9 Macro F1) and the best model with retrieval (\texttt{longformer} with 85.5 Macro F1) is significant ($p<0.001$, $d=5.471$).

\section{Conclusion}
\label{sec:conclusion}

This paper introduces a novel method for fallacy detection and classification relying on a RAG approach to leverage both external knowledge and argumentative structural knowledge. Through the constitution of a 15 GB KB assembled around the entities, topics and time span of the \texttt{ElecDeb60to20} corpus, we evaluated 14 models across 42 retrieval configurations per task, showing that our method outperforms no-retrieval and basic-retrieval baselines. 
On \textbf{fallacy detection}, retrieval improves every encoder, raising the best Macro F1 from $0.772$ to $0.864$ ($p < 10^{-4}$, $d = 8.57$): adding the relations to the query is worth $6.3$ Macro F1 points on average.
On \textbf{fallacy classification} the performance rises from $0.653$ to $0.725$, a significant improvement ($p < 0.01$, $d = 2.07$) that collapses the seed variance of the least stable baselines.

This represents a step toward interpretable fallacy \textit{explanation}: besides predicting fallacy labels, leveraging retrieved context would allow us to explain \textit{why} an argument is fallacious, ensuring transparency in automated political discourse analysis.

\section*{Limitations}
\label{sec:limitations}


Computational cost forced a three-stage protocol in which the 42 retrieval configurations were explored using only the three strongest baseline models per task, and the winning configuration was then applied to all 14. The chosen configuration is therefore optimal for those three models and merely reasonable for the rest, whose optimum may lie elsewhere in the space.

All experiments use one corpus of English-language U.S. presidential debates from 1960 to 2020, and the KB is constructed specifically around it, drawing on U.S. legislative, census and congressional-biography sources. We applied our methodology to a domain close to ours, propaganda detection, with state-of-the-art results. However, a complete shift in the domain would require collecting a different KB.

\section*{Acknowledgments}
This work was supported by the French government through the 3IA Côte d'Azur Investments in the Future project, managed by the National Research Agency (ANR) under reference ANR-23-IACL-0001, and through the France 2030 investment plan, managed by the ANR as part of the Université Côte d'Azur Initiative of Excellence under reference ANR-15-IDEX-01. This work is also supported by ERC grant PANDORA – 101231869. Funded by the European Union. Views and opinions expressed are however those of the author(s) only and do not necessarily reflect those of the European Union or the European Research Council Executive Agency. Neither the European Union nor the granting authority can be held responsible for them. This work was granted access to the HPC/AI resources of IDRIS under the allocation 2026-AD011017429 made by GENCI, on the A100 partition of the Jean Zay supercomputer. The authors also thank Franck Diard for providing access to the DR-1 GPU cluster. 
\bibliography{custom}

\appendix
\section{Hyper-parameters}
Table~\ref{tab:combined-hyperparameters} lists the hyperparameters selected for each model, by Macro F1, in both the Baseline and the RAG settings. They cover fallacy detection and classification on \texttt{ElecDeb60to20} and propaganda detection on \texttt{PROPAGANDA}. These are the configurations behind the results in Tables~\ref{tab:results-fallacy-combined} and~\ref{tab:propaganda-results}.

\begin{table*}[th]
\centering
\small
\setlength{\tabcolsep}{1.8pt}
\newcommand{\NA}{\multicolumn{3}{c}{--}}
\begin{tabular}{llcccccccccccccccccc}
\toprule
& & \multicolumn{6}{c}{\textbf{Fallacy Detection}} & \multicolumn{6}{c}{\textbf{Fallacy Classification}} & \multicolumn{6}{c}{\textbf{Propaganda Detection}} \\
\cmidrule(lr){3-8} \cmidrule(lr){9-14} \cmidrule(lr){15-20}
& & \multicolumn{3}{c}{Baseline} & \multicolumn{3}{c}{RAG} & \multicolumn{3}{c}{Baseline} & \multicolumn{3}{c}{RAG} & \multicolumn{3}{c}{Baseline} & \multicolumn{3}{c}{RAG} \\
\cmidrule(lr){3-5} \cmidrule(lr){6-8} \cmidrule(lr){9-11} \cmidrule(lr){12-14} \cmidrule(lr){15-17} \cmidrule(lr){18-20}
\textbf{Model} & \textbf{Set.} & BS & LR & Ep. & BS & LR & Ep. & BS & LR & Ep. & BS & LR & Ep. & BS & LR & Ep. & BS & LR & Ep. \\
\midrule
RoBERTa-base           & finetune & 16 & $2e^{-5}$ & 5 & 8  & $3e^{-5}$ & 5 & 8  & $5e^{-5}$ & 5 & 8  & $3e^{-5}$ & 5 & 8  & $2e^{-5}$ & 3 & 16 & $5e^{-5}$ & 3 \\
DeBERTa-v3-base        & finetune & 8  & $2e^{-5}$ & 5 & 16 & $2e^{-5}$ & 3 & 8  & $2e^{-5}$ & 5 & 16 & $5e^{-5}$ & 5 & 16 & $2e^{-5}$ & 3 & 16 & $2e^{-5}$ & 3 \\
ModernBERT-base        & finetune & 8  & $3e^{-5}$ & 5 & 8  & $5e^{-5}$ & 5 & 8  & $3e^{-5}$ & 5 & 8  & $5e^{-5}$ & 5 & 16 & $2e^{-5}$ & 5 & 8  & $3e^{-5}$ & 5 \\
NeoBERT                & finetune & 8  & $5e^{-5}$ & 5 & 16 & $2e^{-5}$ & 5 & 8  & $3e^{-5}$ & 5 & 16 & $2e^{-5}$ & 5 & 8  & $3e^{-5}$ & 3 & 16 & $5e^{-5}$ & 3 \\
LegalBERT              & finetune & 8  & $5e^{-5}$ & 5 & 16 & $2e^{-5}$ & 5 & 8  & $5e^{-5}$ & 5 & 16 & $5e^{-5}$ & 5 & 16 & $2e^{-5}$ & 3 & 16 & $3e^{-5}$ & 3 \\
Longformer-base-4096   & finetune & 16 & $5e^{-5}$ & 5 & 16 & $5e^{-5}$ & 3 & 8  & $5e^{-5}$ & 5 & 8  & $3e^{-5}$ & 5 & 8  & $2e^{-5}$ & 3 & 8  & $2e^{-5}$ & 3 \\
RooseBERT-scr-cased    & finetune & 16 & $3e^{-5}$ & 5 & 8  & $5e^{-5}$ & 5 & 8  & $5e^{-5}$ & 5 & 8  & $5e^{-5}$ & 5 & 16 & $5e^{-5}$ & 3 & 16 & $5e^{-5}$ & 5 \\
RooseBERT-scr-uncased  & finetune & 8  & $3e^{-5}$ & 5 & 8  & $5e^{-5}$ & 5 & 8  & $5e^{-5}$ & 5 & 16 & $2e^{-5}$ & 5 & 16 & $3e^{-5}$ & 3 & 16 & $5e^{-5}$ & 3 \\
RooseBERT-cont-cased   & finetune & 16 & $3e^{-5}$ & 5 & 8  & $5e^{-5}$ & 5 & 8  & $5e^{-5}$ & 5 & 8  & $5e^{-5}$ & 5 & 8  & $3e^{-5}$ & 3 & 8  & $3e^{-5}$ & 3 \\
RooseBERT-cont-uncased & finetune & 8  & $5e^{-5}$ & 5 & 8  & $3e^{-5}$ & 5 & 16 & $5e^{-5}$ & 5 & 8  & $5e^{-5}$ & 5 & 8  & $2e^{-5}$ & 3 & 8  & $5e^{-5}$ & 3 \\
\midrule
Llama-3.1-8B-Instruct    & zero-shot & \NA & \NA & \NA & \NA & \NA & \NA \\
                         & few-shot & \NA & \NA & \NA & \NA & \NA & \NA \\
                         & finetune & 8 & $5e^{-4}$ & 3 & 8 & $5e^{-4}$ & 3 & 8 & $3e^{-4}$ & 3 & 8 & $1e^{-4}$ & 1 & 8 & $3e^{-4}$ & 3 & 8 & $3e^{-4}$ & 2 \\
\cmidrule(lr){1-20}
Mistral-7B-Instruct-v0.3 & zero-shot & \NA & \NA & \NA & \NA & \NA & \NA \\
                         & few-shot & \NA & \NA & \NA & \NA & \NA & \NA \\
                         & finetune & 8 & $5e^{-4}$ & 3 & 8 & $1e^{-4}$ & 1 & 8 & $3e^{-4}$ & 2 & 8 & $1e^{-4}$ & 1 & 8 & $1e^{-4}$ & 3 & 8 & $3e^{-4}$ & 3 \\
\cmidrule(lr){1-20}
Qwen3.5-9B-Base          & zero-shot & \NA & \NA & \NA & \NA & \NA & \NA \\
                         & few-shot & \NA & \NA & \NA & \NA & \NA & \NA \\
                         & finetune & 8 & $5e^{-4}$ & 3 & 8 & $5e^{-4}$ & 2 & 8 & $1e^{-4}$ & 3 & 8 & $1e^{-4}$ & 1 & 8 & $5e^{-4}$ & 2 & 8 & $1e^{-4}$ & 2 \\
\cmidrule(lr){1-20}
gpt-oss-20b              & zero-shot & \NA & \NA & \NA & \NA & \NA & \NA \\
                         & few-shot & \NA & \NA & \NA & \NA & \NA & \NA \\
                         & finetune & 8 & $5e^{-4}$ & 3 & 8 & $1e^{-4}$ & 3 & 8 & $5e^{-4}$ & 1 & 8 & $1e^{-4}$ & 1 & 8 & $5e^{-4}$ & 1 & 8 & $5e^{-4}$ & 1 \\
\bottomrule
\end{tabular}
\caption{Best hyperparameters for fallacy detection and classification (\texttt{ElecDeb60to20}) and propaganda detection (\texttt{PROPAGANDA}), Baseline vs.\ RAG. BS = batch size, LR = learning rate, Ep.\ = epochs.}
\label{tab:combined-hyperparameters}
\end{table*}

\section{Prompts}
This section reports the system prompts given to the LLMs for fallacy detection and classification on \texttt{ElecDeb60to20} and propaganda detection on \texttt{PROPAGANDA}. The same prompt is used in all three settings (zero-shot, few-shot and finetuning). In the few-shot setting, the prompt is followed by one labelled example per class, drawn from the training set.

\begin{promptbox}[Fallacy Detection - ElecDeb60to20]
\itshape
You are an expert in argumentation theory and fallacy detection. 
Your task is to determine whether a given sentence from a political debate contains a fallacy. 
A fallacy is a deceptive argumentative move that appears valid but violates principles of correct reasoning.
Answer with exactly one word: "Yes" if the sentence contains a fallacy, or "No" if it does not.
\end{promptbox}

\begin{promptbox}[Fallacy Classification - ElecDeb60to20]
\itshape
You are an expert in argumentation theory and fallacy detection.
Your task is to identify which fallacy type a given sentence from a political debate contains.
A fallacy is a deceptive argumentative move that appears valid but violates principles of correct reasoning.
Consider only these six fallacy types:\\
- Ad Hominem: an excessive attack on the opponent's character (e.g. name-calling and labelling, tu quoque / "you did it first", or bias) instead of addressing their argument.\\
- Appeal to Authority: citing an authority or a group's agreement without providing relevant evidence, including appeals to non-experts or to mere popular acceptance.\\
- Appeal to Emotion: loading the argument with emotional language (appeal to pity, appeal to fear, loaded language, or flag waving) to exploit the audience's feelings.\\
- False Cause: mistaking a correlation between two events for a causal relationship.\\
- Slippery Slope: claiming that an act will lead to an unlikely, exaggerated outcome while omitting the intermediate steps.
- Slogan: a brief and striking phrase used to provoke excitement, often reinforced by repetition.\\
Every sentence contains exactly one of these six fallacy types — there is always a fallacy present, so "None" is never a valid answer.
Answer with exactly one fallacy type, using exactly one of those six names, and nothing else.
\end{promptbox}
\begin{promptbox}[Propaganda Detection - Propaganda]
\itshape
You are an expert in propaganda detection and persuasion analysis.
Your task is to determine whether a given sentence from a news article contains a propaganda technique.
A propaganda technique is a persuasive device (such as loaded language, name calling, exaggeration, appeal to fear, or flag waving) used to influence an audience beyond sound reasoning.
Answer with exactly one word: "Yes" if the sentence contains a propaganda technique, or "No" if it does not.
\end{promptbox}

\section{Additional Results}
This section reports two additional experiments that complement the main results. First, we evaluate argumentative relation classification on \texttt{LIARArg} (Appendix~\ref{appendix:liar-resuls}), which shows how reliably encoders and LLMs predict the relations that FallacyRAG uses to guide retrieval. Second, we compare RAG against a simpler alternative that adds the argumentative relations directly to the input, with no retrieval (Appendix~\ref{sec:input-relation-enriched}). This separates the contribution of the relations themselves from that of retrieval.
\subsection{LIARArg Relation Classification}
\label{appendix:liar-resuls}
In Table \ref{tab:liar-results}, we report the results of the task of relation classification on the \texttt{LIARArg} corpus.
\begin{table}[h]
\centering
\small
\setlength{\tabcolsep}{4pt}
\begin{tabular}{llc}
\toprule
\textbf{Model} & \textbf{Setting} & \textbf{Macro F1} \\
\midrule
RoBERTa-base & finetune & $76.0 \pm 0.7$ \\
DeBERTa-v3-base & finetune & $78.6 \pm 0.8$ \\
ModernBERT-base & finetune & $76.4 \pm 1.0$ \\
NeoBERT & finetune & $57.1 \pm 0.7$ \\
LegalBERT & finetune & $74.5 \pm 1.3$ \\
Longformer-base-4096 & finetune & $76.3 \pm 1.2$ \\
RooseBERT-scr-cased & finetune & $73.9 \pm 0.7$ \\
RooseBERT-scr-uncased & finetune & $72.5 \pm 0.8$ \\
RooseBERT-cont-cased & finetune & $74.7 \pm 0.9$ \\
RooseBERT-cont-uncased & finetune & $74.1 \pm 1.2$ \\
\midrule
\multirow{3}{*}{Llama-3.1-8B-Instruct}
 & zero-shot & $30.3$ \\
 & few-shot & $36.0$ \\
 & finetune & $\mathbf{80.5 \pm 0.6}$ \\
\addlinespace
\multirow{3}{*}{Mistral-7B-Instruct-v0.3}
 & zero-shot & $46.5$ \\
 & few-shot & $49.7$ \\
 & finetune & $76.6 \pm 2.9$ \\
\addlinespace
\multirow{3}{*}{Qwen3.5-9B-Base}
 & zero-shot & $0.2$ \\
 & few-shot & $0.0$ \\
 & finetune & $16.3 \pm 14.0$ \\
\addlinespace
\multirow{3}{*}{gpt-oss-20b}
 & zero-shot & $9.7$ \\
 & few-shot & $0.9$ \\
 & finetune & $78.5 \pm 0.8$ \\
\bottomrule
\end{tabular}
\caption{Argument Component Relation Prediction over \texttt{LIARArg} (Macro F1 mean $\pm$ std, 5 seeds). Best model per column in \textbf{bold}.}
\label{tab:liar-results}
\end{table}

\noindent The prompt used for the decoder is the following:
\begin{promptbox}[Relation Classification - LIARArg]
\itshape
You are an expert in argumentation analysis. You are given two sentences drawn from a fact-checking argumentation corpus: a claim (Sentence 1) and a candidate piece of evidence (Sentence 2).
Your task is to identify the argumentative relation that Sentence 2 holds toward Sentence 1. Consider only these three relations:\\
- Support: Sentence 2 provides evidence or reasoning in favour of Sentence 1.\\
- Attack: Sentence 2 provides evidence or reasoning against Sentence 1.\\
- No Relation: Sentence 2 is not argumentatively related to Sentence 1.\\
Answer with exactly one of those three relation names, and nothing else.
\end{promptbox}

\subsection{Relation-Enriched Input}
\label{sec:input-relation-enriched}
To isolate the contribution of argumentative relations from that of retrieval, we tested an additional setting in which the relations are provided directly in the model's input for the encoders and in the prompt for the decoders, with no retrieval.
\begin{table*}[h]
\centering
\resizebox{\textwidth}{!}{%
\begin{tabular}{llcccccc}
\toprule
& & \multicolumn{3}{c}{\textbf{Detection}} & \multicolumn{3}{c}{\textbf{Classification}} \\
\cmidrule(lr){3-5} \cmidrule(lr){6-8}
\textbf{Model} & \textbf{Setting} & Baseline & Input Enriched & RAG & Baseline & Input Enriched & RAG \\
\midrule
RoBERTa-base             & finetune & 74.7 $\pm$ 0.8 & 76.2 $\pm$ 1.7 & \textbf{80.9 $\pm$ 1.5} & 34.5 $\pm$ 25.0 & 55.9 $\pm$ 4.0 & \textbf{68.0 $\pm$ 2.3} \\
DeBERTa-v3-base          & finetune & 74.2 $\pm$ 6.1 & 76.1 $\pm$ 1.5 & \textbf{77.5 $\pm$ 2.2} & 43.5 $\pm$ 13.0 & 36.5 $\pm$ 15.9 & \textbf{45.5 $\pm$ 19.7} \\
ModernBERT-base          & finetune & 76.4 $\pm$ 1.7 & 71.9 $\pm$ 1.3 & \textbf{86.4 $\pm$ 1.2} & 57.7 $\pm$ 2.1 & 57.7 $\pm$ 6.0 & \textbf{72.5 $\pm$ 4.0} \\
NeoBERT                  & finetune & \textbf{57.2 $\pm$ 21.9} & 40.4 $\pm$ 15.7 & 50.8 $\pm$ 24.0 & \textbf{24.9 $\pm$ 6.8} & 22.8 $\pm$ 2.3 & 10.8 $\pm$ 9.8 \\
LegalBERT                & finetune & 71.7 $\pm$ 1.5 & 72.6 $\pm$ 0.8 & \textbf{77.4 $\pm$ 1.7} & 53.9 $\pm$ 23.7 & 48.0 $\pm$ 11.7 & \textbf{60.8 $\pm$ 3.7} \\
Longformer-base-4096     & finetune & 74.5 $\pm$ 2.0 & 74.4 $\pm$ 1.9 & \textbf{78.7 $\pm$ 1.9} & \textbf{63.3 $\pm$ 2.0} & 57.9 $\pm$ 2.0 & 58.2 $\pm$ 4.8 \\
RooseBERT-scr-cased      & finetune & 77.2 $\pm$ 1.2 & 73.4 $\pm$ 0.6 & \textbf{78.8 $\pm$ 1.1} & 63.0 $\pm$ 2.4 & 61.8 $\pm$ 3.8 & \textbf{69.1 $\pm$ 2.2} \\
RooseBERT-scr-uncased    & finetune & 69.5 $\pm$ 0.9 & 68.3 $\pm$ 0.9 & \textbf{74.3 $\pm$ 1.8} & \textbf{40.1 $\pm$ 4.6} & 34.8 $\pm$ 4.6 & 27.0 $\pm$ 2.7 \\
RooseBERT-cont-cased     & finetune & 74.7 $\pm$ 2.3 & 74.2 $\pm$ 2.1 & \textbf{77.9 $\pm$ 1.1} & 65.3 $\pm$ 4.3 & 59.6 $\pm$ 2.4 & \textbf{68.8 $\pm$ 1.5} \\
RooseBERT-cont-uncased   & finetune & 74.2 $\pm$ 2.5 & 73.1 $\pm$ 1.9 & \textbf{77.0 $\pm$ 2.7} & 44.2 $\pm$ 12.4 & 41.4 $\pm$ 4.3 & \textbf{54.6 $\pm$ 24.4} \\
\midrule
\multirow{3}{*}{Llama-3.1-8B-Instruct}
 & zero-shot & 56.4 $\pm$ 0.0 & \textbf{60.7 $\pm$ 0.0} & 54.3 $\pm$ 0.0 & \textbf{36.4 $\pm$ 0.0} & 27.2 $\pm$ 0.0 & 22.7 $\pm$ 0.0 \\
 & few-shot  & 59.0 $\pm$ 0.0 & \textbf{62.1 $\pm$ 0.0} & 33.4 $\pm$ 0.0 & \textbf{3.6 $\pm$ 0.0} & 0.0 $\pm$ 0.0 & 0.0 $\pm$ 0.0 \\
 & finetune  & 70.7 $\pm$ 1.2 & \textbf{71.3 $\pm$ 1.1} & 63.0 $\pm$ 2.3 & \textbf{53.3 $\pm$ 1.9} & 42.8 $\pm$ 3.8 & 20.2 $\pm$ 1.7 \\
\addlinespace
\multirow{3}{*}{Mistral-7B-Instruct-v0.3}
 & zero-shot & \textbf{56.1 $\pm$ 0.0} & 54.7 $\pm$ 0.0 & 41.1 $\pm$ 0.0 & \textbf{31.5 $\pm$ 0.0} & 22.6 $\pm$ 0.0 & 19.7 $\pm$ 0.0 \\
 & few-shot  & 59.1 $\pm$ 0.0 & \textbf{59.3 $\pm$ 0.0} & 33.4 $\pm$ 0.0 & 1.6 $\pm$ 0.0 & 0.0 $\pm$ 0.0 & \textbf{3.6 $\pm$ 0.0} \\
 & finetune  & \textbf{57.5 $\pm$ 8.3} & 46.9 $\pm$ 5.0 & 47.0 $\pm$ 3.3 & 34.3 $\pm$ 12.9 & \textbf{38.0 $\pm$ 4.7} & 19.1 $\pm$ 4.3 \\
\addlinespace
\multirow{3}{*}{Qwen3.5-9B-Base}
 & zero-shot & \textbf{33.4 $\pm$ 0.0} & 33.3 $\pm$ 0.0 & \textbf{33.4 $\pm$ 0.0} & \textbf{1.6 $\pm$ 0.0} & 0.0 $\pm$ 0.0 & 1.2 $\pm$ 0.0 \\
 & few-shot  & \textbf{33.4 $\pm$ 0.0} & 33.3 $\pm$ 0.0 & \textbf{33.4 $\pm$ 0.0} & 0.0 $\pm$ 0.0 & 0.0 $\pm$ 0.0 & 0.0 $\pm$ 0.0 \\
 & finetune  & 33.5 $\pm$ 0.2 & \textbf{45.9 $\pm$ 18.0} & 33.4 $\pm$ 0.0 & \textbf{1.3 $\pm$ 1.9} & 0.2 $\pm$ 0.5 & 0.0 $\pm$ 0.0 \\
\addlinespace
\multirow{3}{*}{gpt-oss-20b}
 & zero-shot & \textbf{33.4 $\pm$ 0.0} & 33.3 $\pm$ 0.0 & \textbf{33.4 $\pm$ 0.0} & 0.0 $\pm$ 0.0 & 0.0 $\pm$ 0.0 & 0.0 $\pm$ 0.0 \\
 & few-shot  & \textbf{33.4 $\pm$ 0.0} & 33.3 $\pm$ 0.0 & \textbf{33.4 $\pm$ 0.0} & 0.0 $\pm$ 0.0 & 0.0 $\pm$ 0.0 & 0.0 $\pm$ 0.0 \\
 & finetune  & 63.8 $\pm$ 2.5 & \textbf{64.7 $\pm$ 3.9} & 54.0 $\pm$ 3.3 & 34.3 $\pm$ 7.4 & \textbf{35.3 $\pm$ 4.0} & 13.4 $\pm$ 3.9 \\
\bottomrule
\end{tabular}%
}
\caption{Macro F1 (mean $\pm$ std, 5 seeds) on ElecDeb60to20 with baseline, relation-enriched and RAG input. RAG uses a Sentence-RooseBERT retriever with hybrid search (temporal context for detection, none for classification). Best result per model and task in \textbf{bold}.}
\label{tab:elecdeb-sentence-enriched-results}
\end{table*}

As shown in Table~\ref{tab:elecdeb-sentence-enriched-results}, adding the relations directly to the input is never the best option for encoders in either task. Gains over the baseline are small and inconsistent (e.g., +1.5 for RoBERTa in detection), and most models lose performance, especially in classification. Decoders benefit slightly more, mainly in detection: Llama-3.1-8B gains in all three settings (up to +4.3 in zero-shot). Still, they stay well below the encoders: the best LLM with enriched input (Llama-3.1-8B, finetune, 71.3 Macro F1) trails the best encoder with RAG (ModernBERT, 86.4) by 15 points. Overall, relations work better as a signal to guide retrieval than as extra input text, at least for encoders, where RAG gives the best detection result for nine of ten models.
\end{document}